\documentclass[10pt,twocolumn,letterpaper]{article}

\usepackage[pagenumbers]{cvpr} 

\usepackage{graphicx}
\usepackage{amsmath}
\usepackage{amssymb}
\usepackage{booktabs}
\usepackage[utf8]{inputenc} 
\usepackage[T1]{fontenc}    
\usepackage{hyperref}       
\usepackage{url}            
\usepackage{amsfonts}       
\usepackage{nicefrac}       
\usepackage{microtype}      
\usepackage{xcolor}         

\usepackage{times}
\usepackage{bbding}
\usepackage{multirow}
\usepackage{paralist, tabularx}
\usepackage{caption}
\usepackage{subcaption}
\usepackage{pifont}

\usepackage{hyperref}
\hypersetup{
    colorlinks=true,
    linkcolor=blue,
    filecolor=magenta,  
    urlcolor=magenta,
    pdftitle={Overleaf Example},
    pdfpagemode=FullScreen,
    }
\usepackage{color,colortbl}
\definecolor{darkgreen}{rgb}{0,0.5,0}

\definecolor{darkgreen}{rgb}{1,0,0}

\definecolor{azureblue}{rgb}{0,0.5,1}

\usepackage[capitalize]{cleveref}
\crefname{section}{Sec.}{Secs.}
\Crefname{section}{Section}{Sections}
\Crefname{table}{Table}{Tables}
\crefname{table}{Tab.}{Tabs.}

\title{Regularity Learning via Explicit Distribution Modeling \\ for Skeletal  Video Anomaly Detection}

\author{
    Shoubin Yu\textsuperscript{\rm 1},
        Zhongyin Zhao\textsuperscript{\rm 1}, 
        Haoshu Fang\textsuperscript{\rm 1},
        Andong Deng\textsuperscript{\rm 1},\\
        Haisheng Su\textsuperscript{\rm 2},
        Dongliang Wang\textsuperscript{\rm 2},
        Weihao Gan\textsuperscript{\rm 2,3},
        Cewu Lu\textsuperscript{\rm 1},
        Wei Wu\textsuperscript{\rm 2,3}
    \\
    \textsuperscript{\rm 1} Shanghai Jiao Tong University, \textsuperscript{\rm 2} SenseTime Research, \textsuperscript{\rm 3}  Shanghai AI Laboratory \\
}

\begin{document}

\maketitle

\begin{abstract}
Anomaly detection in surveillance videos is challenging and important for ensuring public security. Different from pixel-based anomaly detection methods, pose-based methods utilize highly-structured skeleton data, which decreases the computational burden and also avoids the negative impact of background noise. However, unlike pixel-based methods, which could directly exploit explicit motion features such as optical flow, pose-based methods suffer from the lack of alternative dynamic representation. In this paper, a novel Motion Embedder (ME) is proposed to provide a pose motion representation from the probability perspective. Furthermore, a novel task-specific Spatial-Temporal Transformer (STT) is deployed for self-supervised pose sequence reconstruction. These two modules are then integrated into a unified framework for pose regularity learning, which is referred to as Motion Prior Regularity Learner (MoPRL). MoPRL achieves the state-of-the-art performance by an average improvement of 4.7\% AUC on several challenging datasets. Extensive experiments validate the versatility of each proposed module. 
\end{abstract}



\section{Introduction}
Video Anomaly Detection (VAD) \cite{chalapathy2019deep}  is a challenging yet essential task in computer vision, which is to recognize frames with abnormal events in the video, such as criminal activities and traffic accidents. However, the undisputed nature that abnormal events are far more infrequent than the normal ones makes it nearly impossible to utilize conventional discriminative methods for VAD. Thus, self-supervised methods are widely deployed in the field to learn the normal pattern. In this way, detecting anomalies can be viewed as a task of recognizing out-of-distribution samples. 

Pixel-based methods \cite{liu2018ano_pred,ionescu2019object,SSMT,gong2019memorizing,zaheer2020old,cai2021appearance,chang2020clustering} have been extensively studied for VAD. In those methods, intuitive motion representations, such as optical flow and frame gradients, have been exploited to improve the model sensitivity towards dynamics. Recently, thanks to the success of pose estimation algorithms, pose, as a more clean and well-structured data, gradually attracted the attention of researchers \cite{2019Multi,2020Learning,markovitz2020graph}. Since pose is immune from background noise and contains high-level semantics, it is naturally a convenient feature for human-related VAD. Consequently, pose-based method is viewed as a more promising approach and expected to outperform the pixel-based counterpart.
\begin{figure}
\centering
\includegraphics[width=\linewidth]{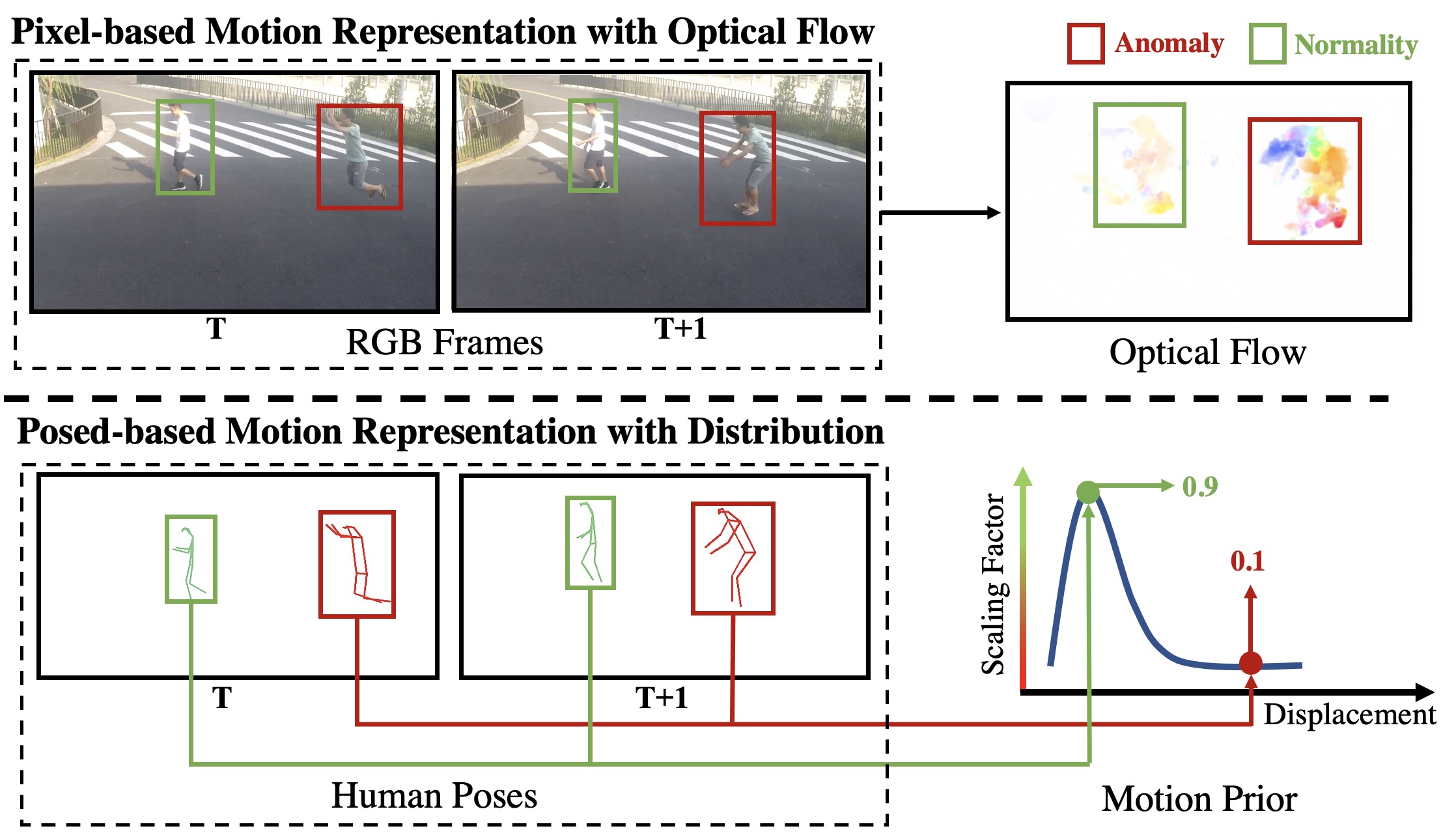}
\caption{In pixel-based methods, optical flow is a commonly-used motion feature; however, the similar feature could not be obtained via pose modality. In this paper, we propose a distribution-based pose motion representation from the perspective of probability. Best viewed in color.}
\label{fig:overview} 
\vspace{-0.3cm}
\end{figure}

Nonetheless, from previous works, even possessing the aforementioned advantages, the performance gain of pose-based methods is still limited. Herein, we claim the reason lies in the fact that, different from image anomaly detection that relies on a static feature only (e.g., appearance), VAD depends more on dynamic features. For example, as shown in the top half of Figure~\ref{fig:overview}, the man in the red box may be judged as normal if only given the right frame, while the confidence of this assertion could be weakened if the left frame is available, since he is jumping on the crosswalk. Therefore, an effective motion representation is essential to regular video pattern learning in VAD. However, in pose-based methods, motion representation, like optical flow, is hard to obtain without RGB information. In previous pose-based methods \cite{2020Learning,markovitz2020graph,2019Multi}, 
researchers usually leverage individual keypoints to encode visual features, which lack intuitive motion representation. In such a situation, the detection model will be overwhelmed to learn both motion and normality simultaneously, which undermines its performance. Hence, it is urgent to find a substitute for pose-based methods. 

In this paper, we propose a novel Motion Prior Regularity Learner (MoPRL) to alleviate the aforementioned limitations in pose-based methods. MoPRL is composed of two sub-modules: Motion Embedder (ME) and Spatial-Temporal Transformer (STT). Specifically, ME is designed to extract the spatial-temporal representation of input poses from the perspective of probabilistic. Inspired by the commonly used frame gradients \cite{yu2020cloze} in pixel-based methods, we model the pose motion based on the displacement between center point of the poses among adjacent frames. However, directly applying the displacement as the motion representation is oversimplified. Given the common assumption that the anomaly rarely happens, we further transform such movement into the probability domain. In detail, we obtain the motion prior, which represents an explicit distribution of displacement on the training data, by statistics. In this way, as shown in the bottom half of Figure~\ref{fig:overview}, to represent corresponding motion, every pose displacement is mapped to a certain probability based on the motion prior. Equipped with a designed pose masking strategy, STT is then deployed as a task-specific model to learn the regular patterns with the input of poses and their motion features from ME. Different from previous RNN-based or CNN-based frameworks, transformer is adopted for its self-supervised and sequential input structure, which naturally fit the pose regularity learning. In summary, the main contributions of this paper are three folds:

\setdefaultleftmargin{1em}{1em}{}{}{}{}
\begin{compactitem}
\item Motion Embedder is proposed to intuitively represent pose motion in the probability domain, which provides effective pose motion representation for its regularity learning.
\item Spatial-Temporal Transformer with pose masking and divided attention is further deployed to model regularity in pose trajectories. To the best of our knowledge, it is the first time that transformer has been applied to video anomaly detection.
\item The proposed MoPRL, which consists of Motion Embedder and Spatial-Temporal Transformer, achieves the state-of-the-art performance on two most challenging datasets. Ablation studies demonstrate the effectiveness of each module, and the insight for future work based on the failure case analysis is provided.
\end{compactitem}

\section{Related Works}

\subsection{Self-supervised Video Anomaly Detection}
In self-supervised video anomaly detection, anomalies are recognized as outliers of the distribution of normality. The pixel-based framework handles the problem by reconstruction and prediction. In \cite{hasan2016learning}, Hasan et al. reconstructed the normal appearance \cite{dalal2005histograms} and motion \cite{dalal2006human} features to learn video regularity. In \cite{luo2017revisit}, the authors leveraged sparsing coding to enforce adjacent frames to be encoded with similar reconstruction coefficients. In \cite{liu2018ano_pred}, Liu et al. proposed a future frame prediction framework with the optical flow \cite{ilg2017flownet} as additional input. In \cite{Nguyen_2019_ICCV}, Nguyen et al. proposed a cross-channel translation framework to learn the coherence between motion and appearance. In a recent work \cite{liu2021hybrid}, Liu et al. exploited Conditional Variational Autoencoder to capture the correlation between the frame and the optical flow. 

Recently, pose-based methods have been popular because of their efficiency and immunity from background noise. In \cite{2020Learning}, the authors proposed a connected RNN for learning pose regularity with decomposed keypoints. In \cite{2019Multi}, Rodrigues et al. deployed a multi-timescale prediction framework to model trajectories. Moreover, Markovitz1 et al.\cite{markovitz2020graph} learned poses graph embeddings with autoencoders and generated soft-assignments via clustering. However, it is rather difficult to obtain a sophisticated dynamic feature in pose-based methods due to the lack of RGB information. In this work, we proposed a novel Motion Embedder to generate pose-based motion representation from probability domain.

\subsection{Vision Transformer}
Transformer \cite{vaswani2017attention} has gradually become a mainstream framework for computer vision community \cite{lin2019tsm,wang2016temporal,ren2015faster,su2018cascaded,qing2021temporal} for its tremendous potential in sequence modeling. It has achieved competitive even superior performance compared with CNN-based methods in image classification \cite{dosovitskiy2020vit,liu2021Swin,pmlr-v139-touvron21a}, object detection\cite{2020End}, semantic segmentation\cite{zheng2021rethinking}, etc. Dosovitskiy et al.\cite{dosovitskiy2020vit} viewed an image as a patch sequence and constructed ViT to achieve effective recognition. Carion et al.\cite{2020End} regarded object detection problem as a set direction prediction task and solve it with a transformer-based encoder-decoder called DETR. Similarly, SETR\cite{zheng2021rethinking} is proposed for image context modeling in semantic segmentation. Transformer is also deployed to estimate 3D human pose as in \cite{zheng20213d}. The authors build a divided temporal-spatial transformer to model the pose sequence. HOT-Net \cite{huang2020hot} fully exploits the correlation between joints and object corners to obtain more accurate estimation. In this work, we also utilize a divided transformer similar with \cite{zheng20213d} to process the scaled pose data obtained from our former Motion Embedder.

\section{Methods}
\begin{figure}
  \centering
\setlength{\abovecaptionskip}{-0.cm} 
  \includegraphics[width=\linewidth]{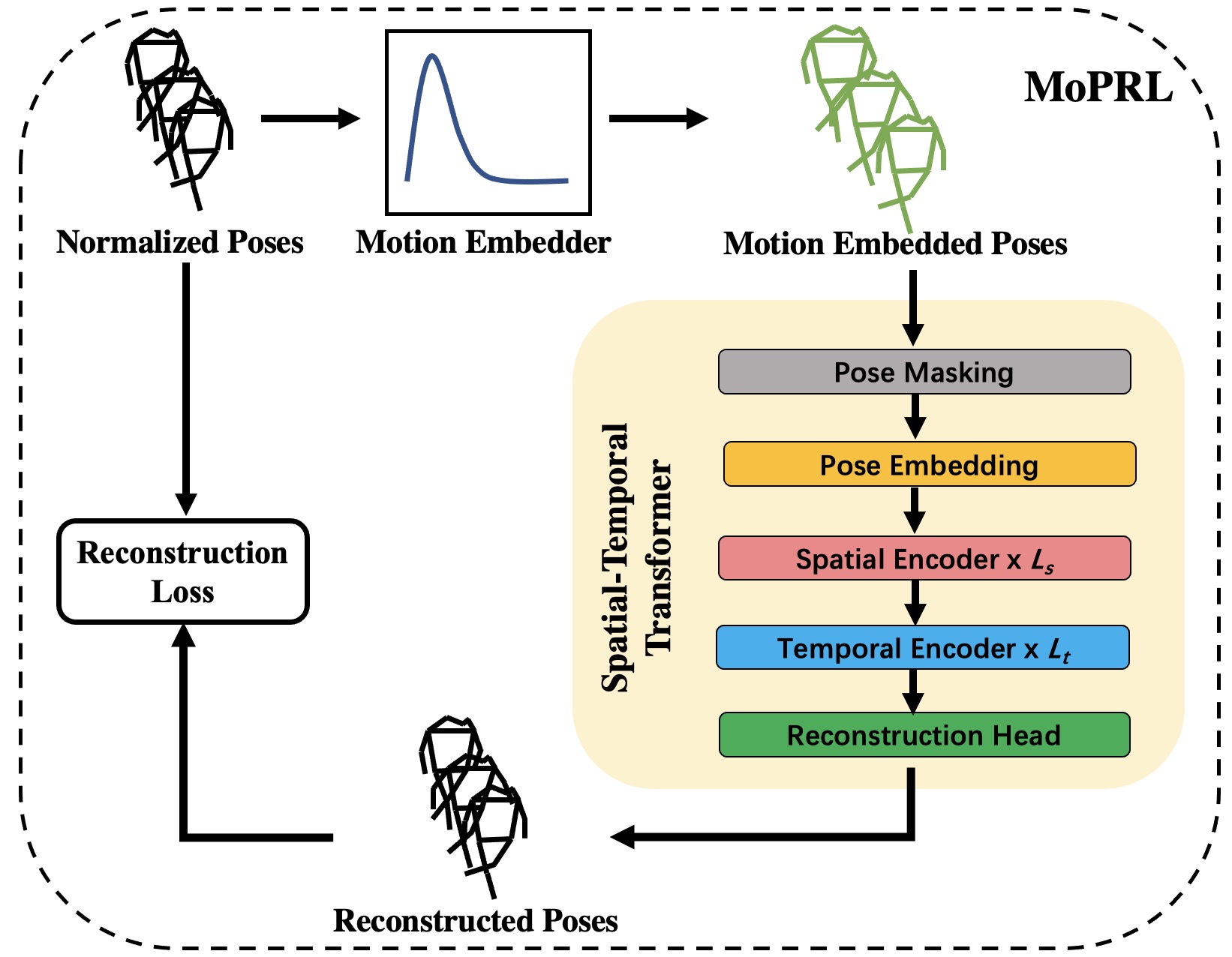}
  \caption{Overview of the proposed Motion Prior Regularity Learner (MoPRL). MoPRL has two sub-modules: one is Motion Embedder, which is used to provide spatial-temporal pose motion representation; the other is Spatial-Temporal Transformer, which is adopted to learn the regularity in motion embedded pose trajectories via reconstruction task. Best viewed in color.}
\label{fig:pipleline} 
\vspace{-0.3cm}
\end{figure}

In this section, we introduce our proposed pose-based video anomaly detection method called Motion Prior Regularity Learner (\textbf{MoPRL}), which consists of two sub-modules as shown in Figure~\ref{fig:pipleline}, namely Motion Embedder (\textbf{ME}) and Spatial-Temporal Transformer (\textbf{STT}). We first utilize a pose detector to obtain the pose trajectories. Unlike pixel-based methods, which adopt the widely-used optical flow as the motion representation, MoPRL models the posed-based motion representation as a probability distribution according to the statistical velocity and fuses spatial and temporal representation via the ME. Then, STT is applied to learn the spatial-temporal regularity with a self-supervised reconstruction task. In this way, the model learns the distribution of the normal samples; thus, the anomalies could be detected according to the Frame Anomaly Score, which will be discussed in detail.

\subsection{Task Definition}
\label{sec:definition} 
Given the training set $D_{train}$ = \{$F_1$, ..., $F_m$\} and the test set $D_{test}$ = \{($F_1$, $L_1$), ..., ($F_n$, $L_n$)\}, where $F_i$ represents the frames and $L_i \in$ \{0, 1\} indicates the label of normality or anomaly. There are only normal samples in training set and both normal and abnormal ones in test set. We denote $F_i$ = \{$S_1$, ..., $S_l$\} and $S_i$=\{$P_1$, ..., $P_t$ \}, which means each frame contains $l$ trajectory sequences of human pose and each sequence consists of $t$ single poses. We denote $P_i$ = \{$J_{i,1}$, ..., $J_{i,k}$\}, where $J_{i,j}$ means the $j$-th joint in the $i$-th pose, and $k$ represents the maximum number of joints in single pose. Moreover, joint $J_{i,j}$ is represented as a coordinate $({x}_{i,j}, {y}_{i,j})$.  A sliding window strategy is applied to match a single frame with its corresponding trajectories. The goal of pose-based video anomaly detection is to distinguish the anomaly with $L_i=1$ in the test set according to those human poses. 
\subsection{Pose Pre-processing}
\label{sec:prepare} 
Following the prepossessing operation proposed in \cite{2020Learning}, we decompose the original pose into a local normalized pose and a global center point. We first calculate the center point $(\Tilde{x}_{i}, \Tilde{y}_{i})$ and the size $(w_{i}, h_{i})$ of the human box according to the maximum and minimum coordinates of the keypoints, and then normalize the pose as $\overline{P}_i=[\overline{J}_{i,1},...,\overline{J}_{i,k}]$ based on the human box size, where $\overline{J}_{i,j}=(\overline{x}_{i,j},\overline{y}_{i,j})$ is the normalized coordinates. The normalized pose $\overline{P}_i$ unifies the scale in different distances, so even tiny changes from far poses can be amplified and captured.
\subsection{Motion Embedder}
\label{sec:ME} 
Since dense motion features, such as optical flow, cannot be obtained, pose-based methods are lack of effective motion representations. In this work, we propose a multi-step approach to get the intuitive pose motion and embed it with pose via the novel Motion Embedder (ME). We first calculate normalized displacement between adjacent poses in sequence and then obtain an explicit discrete distribution describing the training dataset displacement statistic. After this, we choose a predefined distribution (e.g., Rayleigh or Gaussian) to fit the discretized distribution and obtain its continuous version, which we refer to as motion prior. In the end, we leverage both the normalized pose and its motion probability, which represent spatial and temporal information, respectively, to obtain the motion embedded pose. We will introduce ME in detail in the following subsections.
\begin{figure}
  \centering
  \setlength{\abovecaptionskip}{-0.cm} 
  \includegraphics[width=0.8\linewidth]{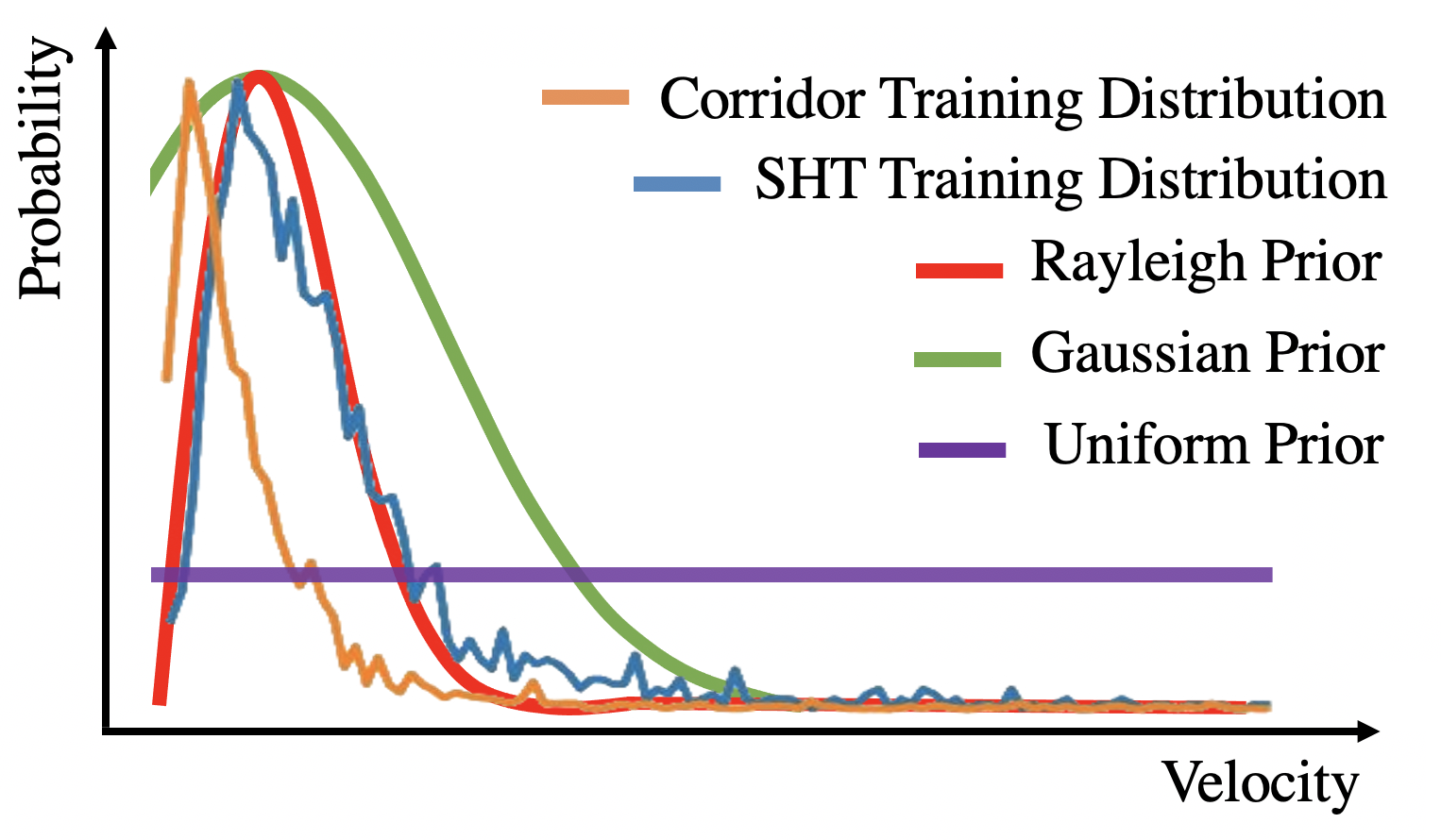}
  \caption{Comparisons of several typical priors with the pose displacement statistic on the training set of two datasets.}
\label{fig:distribution} 
\vspace{-0.2cm}
\end{figure}

\noindent\textbf{Displacement Calculation.} 
The displacement between each pose could be regarded as an average velocity during a short period. Thus, we consider utilizing displacement to construct the foundation of motion representation. Empirically, the displacement is calculated as follows:
\begin{equation}
\begin{footnotesize}
v_{i} = \sqrt{(\Tilde{x}_{i+1} - \Tilde{x}_{i})^2 + (\Tilde{y}_{i+1} - \Tilde{y}_{i})^2 } 
\end{footnotesize}
\end{equation}
\begin{equation}
\begin{footnotesize}
\overline{v}_i = \frac{v_i}{(w_i+h_i)}
\end{footnotesize}
\end{equation}
where $v_i$ represents the pose displacement between adjacent frames, which is also the average velocity from pose $P_i$ to pose $P_{i+1}$. Similar to what we have done to pose, we also normalize the velocity to obtain a normalized version $\overline{v}_i$ which eliminates the influence of perspectives. 
Nonetheless, directly leveraging $\overline{v}_i$ as the motion representation is oversimplified that the essence of a normal motion (normality is common) would be overlooked, leading to that the spatial-temporal feature cannot be effectively represented. We resolve this issue from the perspective of the probability. 

\begin{figure}
  \centering
\setlength{\abovecaptionskip}{-0.cm} 
  \includegraphics[width=\linewidth]{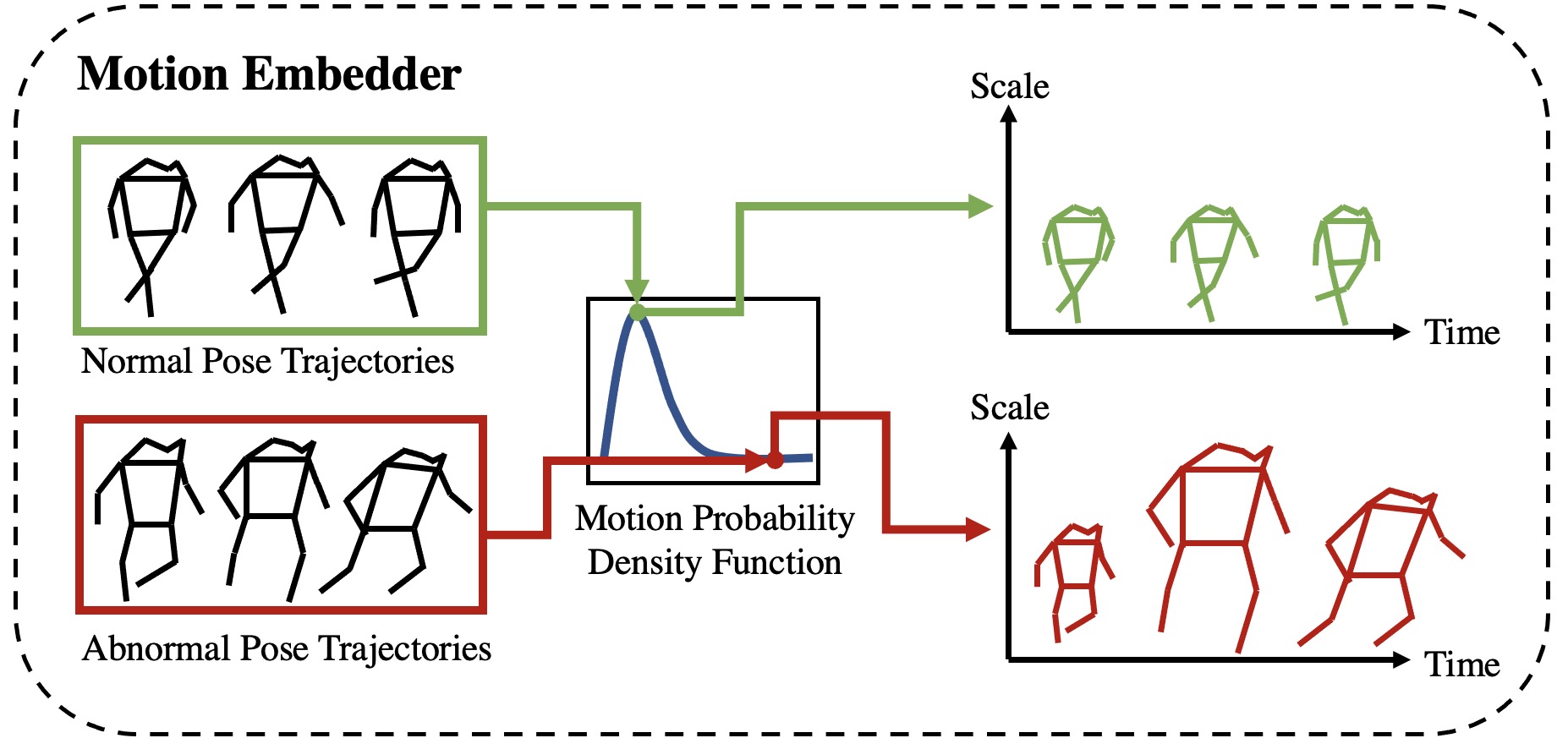}
  \caption{Motion Embedder translates pose motion into the probability domain and embeds motion features into pose appearance. Best viewed in color.}
\label{fig:motion_embedder} 
\vspace{-0.2cm}
\end{figure}

\noindent\textbf{Probabilities as Scaling Factors.}
We first obtain the statistic of the displacements by counting the modes of their normalized version in the training set. Intuitively, we utilize a predefined distribution function to fit this discretized data distribution, by means of which we obtain a continuous displacement distribution. We call this continuous distribution as \textbf{Motion Prior}. Based on the fact that different distribution models its low-frequency part in different manners, we should carefully select an appropriate prior to ensure that it fits more with the real-world distribution, which is believed to be beneficial for the quality of the representation derived from Motion Embedder. As shown in Figure~\ref{fig:distribution}, we can tell that the real distribution of the displacement corresponds more to the Rayleigh distribution, and then is the Gaussian. The experimental results also demonstrate that Rayleigh prior has the best performance. In order to obtain a versatile representation that contains both temporal and spatial information, we expect to combine the normalized pose, which stands for spatial information, and the motion prior, which is actually the temporal representation, we consider employing the probability in motion prior as a scaling factor:
\begin{equation}
\begin{footnotesize}
s_{i} = \rho(\overline{v}_{i}),
\end{footnotesize}
\end{equation}
where $\rho$ is the selected prior to fit the discretized explicit distribution of displacement statistic, and the scaling mechanism is as follows:
\begin{equation}
\begin{footnotesize}
\Hat{P}_{i} = \frac{\overline{P}_i}{s_{i}},
\end{footnotesize}
\end{equation}
where $\Hat{P}_{i}=[\Hat{J}_{i,1},...,\Hat{J}_{i,k}]$ is the pose feature after the scaling operation and represents motion embedded pose that fuses the spatial and temporal information for the $i$-th pose. This is exactly the reason that we call this module Motion Embedder. It is worth noting that, to avoid numerical error, we additionally deploy an affine transformation to the scaling factor, for it may be used as a denominator. Consequently, as shown in Figure~\ref{fig:motion_embedder}, we can obtain a pose with a larger size if the emergence frequency is lower. $\Hat{P}_{i}$ will then be used as the input of the following module.
\subsection{Spatial-Temporal Transformer}
\label{sec:STT} 
To learn the regularity of human pose trajectories, we proposed to utilize transformer-based module to process the motion embedding aforementioned, because of its acknowledged advantage of modeling sequential data. However, the orthodox transformer model results in a computational complexity of $O((N \times T)^{2})$ (where $N$ is the number of joints in a single pose, $T$ the pose number in a single trajectory) and grows exponentially with the increasing $N$ and $T$. Thus, inspired by \cite{gberta_2021_ICML, zheng20213d}, we divided attention mechanism into spatial and temporal parts to decrease the computational complexity to $O(N^2+T^2)$. We call this variant of transformer Spatial-Temporal Transformer (STT). Specifically, STT contains an $L_s$-layer spatial transformer and an $L_t$-layer temporal transformer. Aiming to fully exploit the potential of STT, we view $L_s$ and $L_t$ as hyperparameters and experimentally identify their value, which will be shown in experiment section. In following subsections, we introduce the structure of STT.

\noindent\textbf{Masked Pose Embedding}. Before the transformer blocks, we first obtain the embedding of joints. Specifically, for joint $\Hat{J}_{i,j}$, we follow the mask operation proposed in \cite{devlin2018bert} to obtain its masked version, and then we map it into the embedding space to obtain the joint vector $z_{i,j} \in \mathbb{R}^{C}$, where $C$ is the embedding dimension, as the following equation:
\begin{equation}
\begin{footnotesize}
z_{i,j} = E \cdot mask(\Hat{J}_{i,j}) + E^j_{spe},
\end{footnotesize}
\end{equation}
where $mask(\cdot)$ is the mask function that operated on $\Hat{J}_{i,j}$ with a certain probability, $E \in \mathbb{R}^{C \times 2}$ the learnable embedding matrix. Moreover, $E^j_{spe}\in \mathbb{R}^{C}$ represents a learnable spatial position embedding (SPE) deployed to encode the spatial position of the $j$-th joint in a single pose. Notably, the mask operation only works during the training. After this, we obtain the embedding of the $i$-th pose as $Z_i = [z_{i,0},...,z_{i,N}]$. As a result, the embedding matrix for a whole trajectory is $Z = [Z_1,...,Z_T]$.

\noindent\textbf{Spatial Transformer}. We manage to model the trajectory $Z \in \mathbb{R}^{T \times N \times C}$ on spatial domain with a $L_s$-layer Spatial Transformer. To be noticed, we conduct self-attention on dimension of joint number, i.e., $N$.
Without loss of generality, we denote the input trajectory of the $l$-th layer as $Z^l$, where $l \in [1, L_s]$. The multi-layer attention operation is given by:
\begin{equation}
\label{qkv}
\begin{footnotesize}
Q=Z^l_{ln} \cdot W_{Q}, K=Z^l_{ln}\cdot W_{K}, V=Z^l_{ln}\cdot W_{V},
\end{footnotesize}
\end{equation}
\begin{equation}
\label{softmax}
\begin{footnotesize}
\Hat{Z}^{l+1} = softmax(QK^{T}/\sqrt{C})V+Z^l,
\end{footnotesize}
\end{equation}
\begin{equation}
\label{ln}
\begin{footnotesize}
{Z}^{l+1} = LayerNorm(fc(\Hat{Z}^{l+1}_{ln})+\Hat{Z}^{l+1}),
\end{footnotesize}
\end{equation}
where $Q, K, V $ is the query, key and value matrix, $W_{Q}$, $W_{K}$, $W_{V} \in \mathbb{R}^{C \times C}$ the corresponding project heads. The subscript ${ln}$ indicates a tensor after layer normalization. $softmax$ and $fc$ represents softmax operation and fully-connected layer, respectively. Actually, we leverage multi-head self-attention as our attention operation for stronger representation. As it has been a common structure, we ignore its formulation here for simplicity. Please refer to \cite{vaswani2017attention} for more details.

\noindent\textbf{Temporal Transformer}. We then model pose trajectories on the temporal domain with a $L_t$-layer Temporal Transformer. Taking the output $Z^{L_s}$ of Spatial Transformer as input, we first incorporate temporal position information for each joints embedding as follows:
\begin{equation}
\begin{footnotesize}
z = z_{i,j}^{L_s} + E_{tpe}^j,
\end{footnotesize}
\end{equation}
where $z_{i,j}^{L_s}$ represents the $j$-th joints embedding in the $i$-th frame of $Z^{L_s}$, $E_{tpe}$ a learnable temporal position embedding (TPE). Thus, the trajectory embedding matrix can be obtained accordingly. Following the same steps in Equation ~\ref{qkv}, ~\ref{softmax} and ~\ref{ln} in aforementioned spatial transformer, we finally obtain the spatial-temporal output $Z^o$.

\subsection{Self-supervised Training}
\label{sec:training} 
In this subsection, we introduce our self-supervised training to learn the regularity in human pose trajectories. Specifically, we achieve this via the commonly used reconstructive method. With the help of a reconstruction head, we are able to learn the distribution of the normal samples.

\noindent\textbf{Reconstruction Head}. As shown in the left side of Figure~\ref{fig:pipleline}, taking motion embedded trajectory $[\Hat{P}_1,...,\Hat{P}_t]$ as input, the Reconstruction Head recovers the normalized trajectory $\overline{S}=[\overline{P}_1,...,\overline{P}_t]$ from the output of the STT $Z^o$.
\begin{equation}
\begin{footnotesize}
\overline{S}' = Head_{rec} (Z^o)
\end{footnotesize}
\end{equation}

\noindent\textbf{Objective Functions}. The final objective function can be described as follows:
\begin{equation}
\begin{footnotesize}
Loss = \sum_{i=1}^{T}\sum_{j=1}^{N}{\omega^{i,j}\vert\vert \overline{J}'_{i,j}-\overline{J}_{i,j}\vert\vert_2}
\end{footnotesize}
\end{equation}
where $\omega^{i,j}$ is the confidence score of each pose joint coming from pose detector, and $\overline{J}'_{i,j}$ is the reconstructed joint in $\overline{S}'$. Similar to the operation in \cite{2019Multi}, we also normalize the raw joints confidence.
\subsection{Inference}
\label{sec:inference} 
\noindent\textbf{Frame Anomaly Score}. In this section, we introduce the mechanism that the proposed method detects frame-level anomaly with human pose trajectories. Firstly, an anomaly score $A_{m,n}$, where $n$ and $m$ represent the $n$-th trajectory in the $m$-th frame, will be obtained from each pose trajectory via MoPRL. The anomaly score $A_{m,n}$ is the $L_1$ norm of the difference between $\overline{S}'$ and $\overline{S}$, given by:
\begin{equation}
\begin{footnotesize}
A_{m,n} = \vert\vert{\overline{S}'-\overline{S}}\vert\vert_1.
\end{footnotesize}
\end{equation}
Since each frame may contain multiple trajectories, we select the highest $A_{m,n}$ as the frame-level anomaly score $A_m$:
\begin{equation}
\begin{footnotesize}
A_{m} = Max(A_{m,n})
\end{footnotesize}
\end{equation}
\label{eq:socre}
The higher frame anomaly score $A_m$ suggests higher possibility for the current frame to be abnormal.

\section{Experiments}
In this section, we first introduce experimental details of proposed MoPRL. Extensive experiments are then conducted on two challenging datasets to evaluate the effectiveness and superiority of MoPRL with convincing qualitative examples.
\vspace{-5mm}
\begin{table}
\centering
\caption{Frame-level AUC performance comparisons with other Pose-based methods on two popular datasets. SHT-HR is a subset of SHT, which only contains human-related samples.} 
\vspace{-0.3cm}
\begin{tabular}{c|ccc}
\toprule[1pt]
\textbf{Methods}  & \textbf{SHT} & \textbf{SHT-HR} & \textbf{Corridor}\\ \hline
{MPED-RNN \cite{2020Learning}} & 73.40                 & 75.40  & 64.27 \\               
{GEPC \cite{markovitz2020graph}}     & 75.50                 & \textbackslash{}   & \textbackslash{} \\  
{MTP \cite{2019Multi}}      & 76.03                 & 77.04   & 67.12 \\    \hline          
\textbf{Ours}     & \textbf{81.26}                 & \textbf{82.38}  & \textbf{70.66} \\
\bottomrule[1pt]
\end{tabular}
\label{tab:main_tale}
\end{table}

\begin{table*}
\centering
\caption{Comparison with~\cite{2020Learning} and study of each proposed module on ShanghaiTech. ME represents the Motion Embedder, and STT is the Spatial-Temporal Transformer. $*$: the results are reproduced by us. $\dagger$: only reconstruction task with normalized pose is applied.}
\vspace{-0.2cm}
\begin{tabular}{c|c|c|c|c}
\toprule[1pt]
\textbf{Methods}                          & \textbf{Appearance Representation} & \textbf{Motion Representation} & \textbf{Sequence Modeling} & \textbf{AUC}    \\ \hline
\multirow{2}{*}{MPED-RCNN$\dagger$~\cite{2020Learning}} & \multirow{6}{*}{Pose Embedding}               & \XSolidBrush                                              & \multirow{2}{*}{RNN \cite{schuster1997bidirectional}}                                        & 72.20$*$ \\ \cline{5-5} 
                                  &              & ME       &                                                             & 73.08$*$ \\ \cline{1-1} \cline{3-5} 
 \multirow{4}{*}{Ours}           &                                                                     & \XSolidBrush                          & \XSolidBrush                                       & 68.32  \\ \cline{5-5} 
                                  &                                                                     & \XSolidBrush                                              & STT                                                         & 68.56  \\ \cline{5-5} 
                                  &                                                                     & ME                                                              & \XSolidBrush                                           & 76.92  \\ \cline{5-5} 
                                  &                & ME                                                              & STT                                                         & \textbf{81.26}  \\ \bottomrule[1pt]
 \end{tabular}
 \label{tab:module-level}
 \end{table*}

\subsection{Datasets and Setup}
\noindent\textbf{Datasets}. We evaluate our method on two most challenging datasets: \textbf{ShanghaiTech} \cite{luo2017revisit} contains 330 training videos and 107 testing ones. It consists of 13 training scenes and 12 testing scenes. And ShanghaiTech-HR \cite{2020Learning} is a subset of ShanghaiTech containing only Human-Related anomaly with 101 testing videos. \textbf{Corridor} \cite{2019Multi}, a recent dataset for video anomaly detection with largest size, contains 10 abnormal classes in a single scene. It includes both single and multiple person anomalies. 

\noindent\textbf{Pose Estimator}. For fair comparison, we adopt the same pose estimator with other compared methods to avoid the variance caused by pose quality. Specifically, we use the tools~\cite{fang2017rmpe, xiu2018poseflow} to obtain the pose trajectories as in \cite{markovitz2020graph} on ShanghaiTech. While for the Corridor dataset, we extract pose trajectories with tools~\cite{openpose,chen2018real} as in~\cite{2019Multi}. Note that each pose joint is provided with a confidence score.

\noindent\textbf{Implementation Details}. We apply AdamW~\cite{kingma2014adam} optimizer with an initial learning rate of $5e-5$ and adopt a warm-up schedule with 1000 steps. Empirically, the layers number of Spatial Transformer and Temporal Transformers are both set to $2$. The batch size is $256$ and the dimension of vector embedding is 128. Each trajectory contains 8 poses ($T=8$) and each pose contains $17$ joints for AlphaPose~\cite{fang2017rmpe} results and $25$ joints for OpenPose~\cite{openpose} results (e.g., $N=17$ or $25$). To obtain the pose sequences, we sample the pose trajectory using sliding windows with window size of $16$ and stride $2$. Following BERT~\cite{devlin2018bert}, the mask ratio of poses is set to $0.15$. We normalize the frame-level anomaly scores in each scenario for final evaluation as in~\cite{Nguyen_2019_ICCV}. And all experiments are conducted on the entire dataset without division by scenarios. More hyperparameters experiment reports are included in Supplementary.

\noindent\textbf{Evaluation Metrics}. Following the conventions, Area Under Curve (AUC) is calculated as the evaluation metric. A higher AUC indicates better anomaly detection performance.
\subsection{Comparison with the State-of-the-Arts}

\begin{figure}
  \centering
  \includegraphics[width=\linewidth]{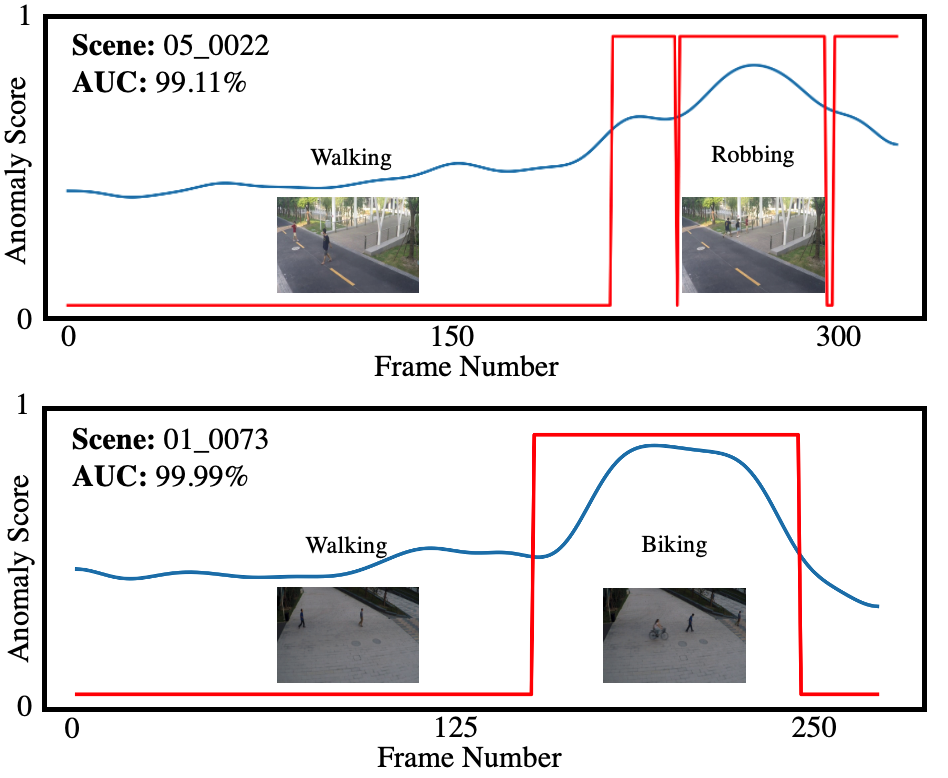}
  \caption{Frame-level anomaly scores (blue lines) corresponding to the labels (red lines) under two scenes in ShanghaiTech dataset. More visualization results are included in Supplementary. Best viewed in color.}
\label{fig:anomaly_score} 
\end{figure}

Table~\ref{tab:main_tale} illustrates the comparison results of our MoPRL with other state-of-the-arts on two popular datasets respectively, namely ShanghaiTech (SHT) and Corridor. We can observe that our method can obtain consistent and significant performance improvement compared with other pose-based methods on all conducted datasets. Qualitative examples are further provided to illustrate the anomaly prediction results of our method, as shown in Figure~\ref{fig:anomaly_score}. Surprisingly, we can find that our method can not only detect obvious anomalous events (e.g., "Biking"), but also is capable of observing some subtle activities, such as "Robbing" happening in a far distance away from the camera. Besides, our MoPRL is sensitive to the the anomalies with extreme movements (e.g., "running" and "pushing") rather than the appearance-based anomaly events (e.g., "holding the suspicious object"), which is reasonable owing to the fact that original RGB information (e.g., clothing and belongings) is absent when only pose modality is adopted as input. However, many object-related anomaly classes (e.g., "wearing a mask" and "carring a box") are included in the Corridor dataset, which accounts for the reason that why performance achieved in ShanghaiTech dataset is higher than the Corridor dataset by a large margin (+10.6\%). Therefore, our proposed method can achieve effective and non-trivial performance in abnormal poses discrimination.

\subsection{Ablation Studies}
In this section, we first conduct a series of exploration studies of each module proposed in our MoPRL method. As shown in Table~\ref{tab:module-level}, a linear auto-encoder (including only an encoder and a decoder) with normalized poses as input is adopted as our baseline, which can only obtain 68.32\% AUC in our experiments. Then the proposed motion embedder is also applied to other pose-based method~\cite{2020Learning} to show the generalization ability and necessity of motion representation.

\noindent\textbf{Motion Representation.} As shown in Table~\ref{tab:module-level}, with the help of ME, significant AUC improvement can be observed even without sequence modeling (+8.60\%), which demonstrates that motion representation is essential for pose-based anomaly detection methods. Without loss of generality, we continue to evaluate the necessity of motion representation in a classical multi-task pose-based method named as MPED-RCNN~\cite{2020Learning}, which adopts RNN for sequence modeling of input pose embedding. To be clarified, we reproduce the MPED-RCNN method to fit the proposed motion prior of ME, and only the reconstruction task with local normalized pose is applied to align the setting of our MoPRL for fair comparison. As expected, the proposed method can bring consistent performance improvement as listed in Table~\ref{tab:module-level}. It confirms that ME can truly benefit other pose-based methods, and the motion representation should be accounted as an essential factor in developing pose-based methods. However, the huge contrast of performance gain in different frameworks (8.60\% vs. 0.88\%) also raises a concern that such a distribution-based hand-crafted feature extractor is still not an optimal way for all pose-based methods. Future research can focus more on how to develop general motion representations for pose modality.



\noindent\textbf{Sequence Modeling.}  We further explore the impact of sequence modeling which is considered as the core of spatial-temporal regularity learning. However, our STT can only bring trivial performance improvement (+0.24\%) without motion representation. After combining with ME, STT can further boost the overall performance by a great margin (+4.34\%), which reveals that the proposed ME module can help to obtain the discriminative motion clues of input poses, and the STT module can further model the temporal inter-dependencies. Thus, we argue that the normalized poses are inferior to describe the motion dynamics of pose trajectories, which hinders the potential of such sequence modeling. We also observe that RNN~\cite{schuster1997bidirectional} adopted in \cite{2020Learning} actually leads to the performance declination (-3.84\%) when ME is applied. It demonstrates that the RNN-based model actually limits the performance gain from such motion representations. We hypothesize that, unlike transformer-based STT benefiting from big data, the cascaded and history-dependent RNN underfits such large data size with limited model capacity. The visualization of reconstructed poses from STT and RNN (please refer to Supplementary) further confirms this assumption. Compared with STT, RNN model reconstructs poses with huge deviation even for normal samples. And such deviation actually shadows the distinction between normality and anomaly brought by the motion prior. Conclusively, both the motion representation and effective sequence modeling are important and indispensable for posed-based VAD.

\subsection{Analysis on Motion Embedder}
\begin{figure}
  \centering
  \includegraphics[width=\linewidth]{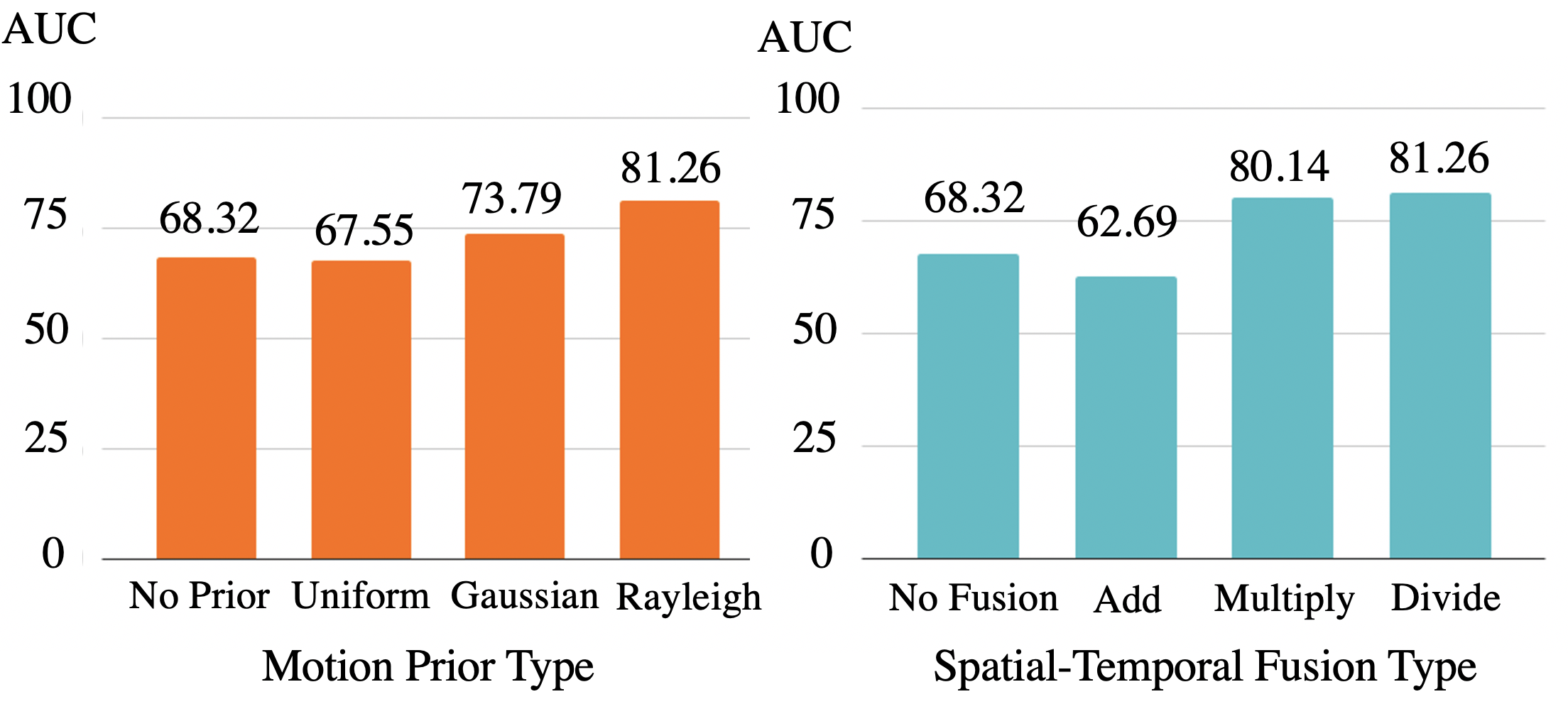}
  \caption{\textbf{Left}: the bar chart of performance comparison among different motion prior distribution types. \textbf{Right}: the chart of performance comparison among different spatial-temporal fusion types. Experiment on ShanghaiTech.}
\label{fig:me_ablation} 
\end{figure}
\noindent\textbf{Motion Prior Type.} As mentioned in Section~\ref{sec:ME}, ME is designed to represent intuitive pose motion by probability prior. Since the distribution of such prior directly controls the probability of pose motion and explicitly decides how rare should the extreme motion be, the selection of prior matters to the final performance. In this section, we chose several explicit priors with obviously different shapes to quantitatively evaluate the impact of the motion prior type. As shown in the left half of Figure ~\ref{fig:me_ablation}, we select uniform prior and Gaussian prior to compare with the Rayleigh prior we applied in our experiments. Referring to Figure~\ref{fig:distribution}, the result shows the larger difference between the selected prior and the statistical distribution on the training dataset is, the more the performance of MoPRL decline (12.07\% with uniform prior and 7.47\% with Gaussian prior). It demonstrates that the model can indeed benefit from of selected motion prior. And the gain will increase if the the discrepancy between the prior and the real-wolrd motion distribution decreases.

\noindent\textbf{Spatial-Temporal Fusion Type.} In order to obtain the spatial-temporal input for STT, ME fuses the motion prior into poses. Hence, the fusion operation mentioned in Section~\ref{sec:ME} should be thoroughly explored to verify its effectiveness. In this section, besides the division, we also deploy other common operation to conduct such fusion. As shown in the right half of Figure ~\ref{fig:me_ablation}, the results show that MoPRL is less sensitive to different scale-related operations (multiply with 80.14\% and division with 81.12\%). While the common fusion strategy in pixel-based methods, e.g., addition, does not work in pose-based methods, impairing the model performance compared with baseline (62.69\% versus 67.58\% ). In this case, poses may just be moved from their original spatial location and perturbed by the fusion. It demonstrates the effectiveness and rationality of our proposed fusion method.

\subsection{Analysis on Spatial-Temporal Transformer} 
\begin{table}
\centering
\caption{Performance and complexity comparison among different attention mechanism. Experiments are conducted on ShanghaiTech.}
\vspace{-0.2cm}
\begin{tabular}{c|cc}
\toprule[1pt]
\textbf{Attention}       & \textbf{Computational Complexity}  & \textbf{AUC} \\ \hline
{No Attention} & \textbackslash{} & 76.92                       \\ 
{Joint} & $(T \times N)^2$ & 77.88                       \\ 
{Spatial Only} & $N^2$ & 78.64                      \\ 
{Temporal Only} & $T^2$ & 80.45                        \\ \hline
{Spatial-Temporal} & $(N^2 + T^2)$ & \textbf{81.26}           \\ 
\bottomrule[1pt]
\end{tabular}
\label{tab:attention}
\end{table}

\noindent\textbf{Attention Mechanism.} A major concern of transformer-based models is its exponentially increasing complexity with input size. Thus, to alleviate this issue, we applied divided spatial and temporal attention. as shown in Table~\ref{tab:attention}, we list comparative results among different attention mechanism for quantitatively evaluation. We establish the baseline as the model without any attention but with motion prior. The joint attention represents the vanilla transformer taking the entire sequence as input. It improves only $0.96\%$ with the most heavy computational burden. Moreover, observing the declination compared with spatial only ($0.76\%$) or temporal only ($2.57\%$) case, we claim that the indifferent attention among the entire sequences would actually harm the performance. Furthermore, the temporal modeling matters more to the performance with the most significant performance boosting ($\sim3\%$), which demonstrates that motion information is essential to the video anomaly detection. Finally, the highest performance with Spatial-Temporal Attention demonstrates the effectiveness of our design.
\begin{table}
\centering
\caption{Study of model depth of spatial and temporal encoders respectively on ShanghaiTech dataset.}
\vspace{-0.2cm}
\begin{tabular}{c|ccccc}
\toprule[1pt]
\textbf{$L_s$ / $L_t$} & \textbf{1}  & \textbf{2}  & \textbf{4} & \textbf{6} & \textbf{8} \\ \hline
\textbf{1}  & 81.02  &81.04 & 81.22 & 80.39 & 80.19\\ 
\textbf{2}  & 81.15  &\textbf{81.26} & 80.14 & 80.86  & 79.42\\ 
\textbf{4} & 75.34  &79.98 & 77.41   & 80.29 & 79.03\\ 
\textbf{6} & 78.25  &74.92 & 78.67  & 79.79  & 78.50 \\ 
\textbf{8}  & 74.25  & 77.34 & 76.94 & 79.28 & 79.07\\ 
\bottomrule[1pt]
\end{tabular}
\label{tab:model_depth}
\end{table}

\noindent\textbf{Model Depth.} In this section, we ablate several combinations of the layer depth in both spatial and temporal dimension. The results listed in Table~\ref{tab:model_depth} demonstrates that MoPRL does not actually benefit from a deeper attention structure. The deepest model brings 2.19\% decrease. Further, compared with the deepest temporal attention which brings an average 2.26\% decrease, the deepest spatial attention leads to a more significant average drop (-3.39\%). 

\begin{figure}
  \centering
  \includegraphics[width=\linewidth]{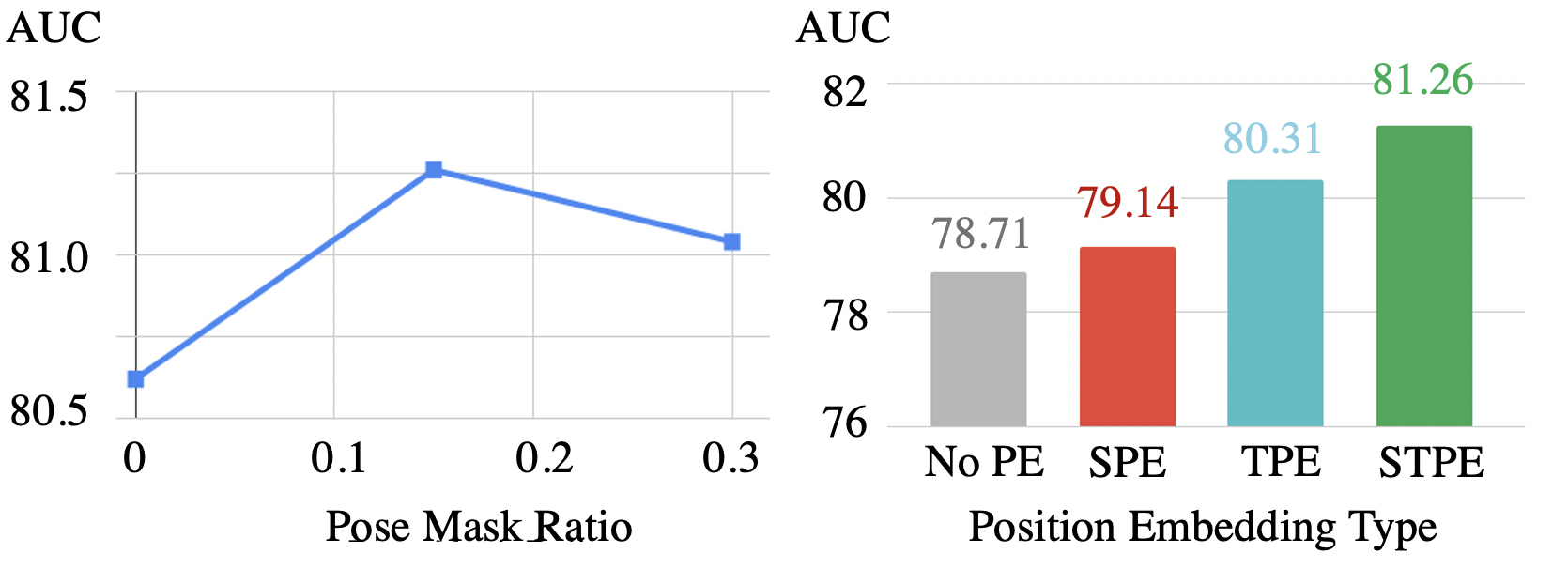}
  \caption{Comparative experiments on ShanghaiTech. \textbf{Left}: the line chart of performance comparison among different mask ratios. \textbf{Right}: the chart of performance comparison among different position embeddings.}
\label{fig:stt_ablation} 
\vspace{-0.2cm}
\end{figure}

\noindent\textbf{Pose Masking and Position Embedding.} In this section, we verify the the effectiveness of pose masking and position embedding. We leverage pose masking as an operation of data augmentation. As shown in the left of Figure~\ref{fig:stt_ablation}, model performance first improves as the pose mask ratio increases and then drops to about 81.0\% but still higher than the performance of the case without pose masking, which certainly demonstrates that model benefits from such perturbation. However, model would fail to handle regularity learning under a too large ratio. Furthermore, we also evaluate the performance change brought by different position embeddings strategies. As shown in the right of Figure~\ref{fig:stt_ablation}, different position embedding approaches utilized on MoPRL bring consistent performance gain, and temporal position embedding contributes more to the model performance.

\subsection{Runtime Analysis}
Running speed is an essential factor for online video anomaly detection. The proposed MoPRL is deployed on an single NVIDIA RTX 3090 GPU. In inference, the best AUC performance setting of MoPRL runs at 108 fps, and the shallowest setting of MoPRL runs at 158 fps with 81.02\% AUC. We also report the running speed of another pose-based method \cite{2020Learning} runs at 161 fps and pixel-based method \cite{liu2021hybrid} runs at about 66 fps on the same GPU. The speed bottleneck of pose-based pipelines lies in the feature extraction, i.e., pose estimation and tracking process, which usually runs at 15 fps. Thus, the end-to-end speed could be $\sim$14 fps.

\section{Discussion}
\noindent\textbf{Failure Cases.}
Pose-based methods depend heavily on the quality of the estimated poses. Therefore, when the pose estimator fails to extract the pose structure, MoPRL would have poor performance (e.g., when the objects are occluded or fast-moving). Besides, the tracking algorithm often captures inaccurate trajectories in a crowd scene, which directly leads to a serious impediment for MoPRl to detect. Moreover, failure cases are also observed in object-relative anomalies (e.g., human with a trailer or a mask, etc.) and motion direction anomalies (e.g., sudden turning around.). The absence of RGB information probably causes this, and Motion Embedder alone cannot capture directional motion feature.

\noindent\textbf{Limitations.} 
Although this work adopts the high-level pose features extracted from the video, we still find it too challenging to only take the displacement for pose motion modeling. Other essential characteristics, like motion direction, are ignored. Thus, in the future, we suggest fully considering all possible motion features to construct a completed and universal pose motion representation for related tasks.

\noindent\textbf{Broader Impacts.} 
The two surveillance video datasets \cite{2019Multi,luo2017revisit} utilized in this work may lead to potential risk of portrait rights, since the frames would inevitably contains unwitting passersby though curated collected.

\section{Conclusion}
In this paper, we propose a novel Motion Prior Regularity Learner (MoPRL) for pose-based video anomaly detection. MoPRL takes pose motion probability from the prior statistics on the training dataset as the intuitive dynamic representation via the proposed Motion Embedder. Then, MoPRL models pose trajectories regularity with the a spatial-temporal transformer equipped with divided attention. It achieves the state-of-the-art on two challenging mainstream datasets. Ablation studies and failure case analysis provide insights for future works.

\newpage
{
\bibliographystyle{abbrv} 
\bibliography{main} 
}



\end{document}